# Improving Word Recognition in Speech Transcriptions by Decision-level Fusion of Stemming and Two-way Phoneme Pruning

Sunakshi Mehra[1[0000-0002-6397-6049]] and Seba Susan[1[0000-0002-6709-6591]]

[1]Department of Information Technology
Shahbad Daulatpur, Main Bawana Road, Delhi 110042, India

**Abstract.** We introduce an unsupervised approach for correcting highly imperfect speech transcriptions based on a decision-level fusion of stemming and two-way phoneme pruning. Transcripts are acquired from videos by extracting audio using Ffmpeg framework and further converting audio to text transcript using Google API. In the benchmark LRW dataset, there are 500 word categories, and 50 videos per class in mp4 format. All videos consist of 29 frames (each 1.16 seconds long) and the word appears in the middle of the video. In our approach we tried to improve the baseline accuracy from 9.34% by using stemming, phoneme extraction, filtering and pruning. After applying the stemming algorithm to the text transcript and evaluating the results, we achieved 23.34% accuracy in word recognition. To convert words to phonemes we used the Carnegie Mellon University (CMU) pronouncing dictionary that provides a phonetic mapping of English words to their pronunciations. A two-way phoneme pruning is proposed that comprises of the two non-sequential steps: 1) filtering and pruning the phonemes containing vowels and plosives 2) filtering and pruning the phonemes containing vowels and fricatives. After obtaining results of stemming and two-way phoneme pruning, we applied decision-level fusion and that led to an improvement of word recognition rate upto 32.96%.

## 1 Introduction

Automatic Speech recognition is the process of transforming speech to recognizable text with the help of computer interface. It is challenging to extract pure text transcript from audio since different speakers may have different styles, different accents, and different voice quality [1]. Speech recognition is not speaker-dependent; recognition of speakers from characteristics of their voices is another line of research [2, 4]. In recent years, a lot of progress has been achieved in the task of spoken text recognition by supervised classification, and less work is focused on unsupervised approaches. Alignment of subset of phonemes on audio track with the sequence of phonemes extracted from the imperfect speech transcript was acknowledged as a challenging task in [5] because of varying sound/video quality compromised by noise disturbances, different speakers, varying accent and varying style in pronunciation of words. The performance of aligning shows a correct matching of phonemes with 10, 20, 30 error margins when more than 75% of text is aligned correctly within 5 seconds. Unaligned temporal text is converted to phonemes and the alignment between audio phonemes and text phonemes is done with dynamic programming edit distance transformation (Haubold and Kender, 2007) [5]. Text to speech synthesizing system based on Grapheme-to-Phoneme conversion resolves problems in rule based approaches [8,13]. The use of root words or sub-words can be observed in [11, 14, 31, 33, 35] in which the prefix and suffix of every word is stripped off. The idea of grapheme-to- phoneme conversion is extended in [10] that uses the sub-word approach in resolving ambiguity in inflected and compound words. The algorithm segments each word into the main part and the suffix part, and concatenates the pronunciations of the two parts procured separately from two different pronunciation dictionaries. The study shows 6% of the errors are achieved due to inflected or compound words among a total of 8.33%. Recently developed online tools like WinPitchPro perform text-to-speech alignment [23]. Unsupervised acoustic modelling on 100k words of text in [25], measures the effects of weak language models by model training. This method uses word confidences to predict n-gram counts for language modelling by multiplying the word confidences together to get a weighted count which results in better reduction in word error rate. A larger gain might be possible but results improve much more in margin if better estimation techniques are used to identify correct n-grams on small amounts of labelled data (Novotney et al., 2009) [25].

Speech separation can also be used for the classification problem, based on supervised learning [17]. Over the past few decades, speech separation and Weiner filtering make the assumption of background noises which constitutes a major problem in our current system. Deep neural network with multiple hidden layers can perform better than Multilayer Perceptron in speech separation (Chen, Wang and Wang, 2014) [18]. Morphological parsing may or may not involve stemming. It may just involve searching for a word in its original form as it appears in the sentence [36, 39]. The stemming framework involves two main aspects: first is to expand the sentence selection strategy from the so-called "sentence by phrase" to "list by phrase", and second is to examine different construction methods and component models involved in such a framework (Lin, Yeh, and Chen, 2010) [15]. The foremost challenge lies in how the word is pronounced and with how much intensity the person said the word. To map the words to phonemes we used the CMU dictionary which is also called a pronouncing dictionary [40]. In this paper, we use a stemming framework in collaboration with phonemes filtered and pruned in two independent stages for recognizing words and correcting errors in imperfect speech transcripts. The organization of this paper is as follows: Section 2 presents the proposed approach, Section 3 analyses the experimental results and Section 4 states the conclusion.

# 2 Proposed Approach for Correcting Imperfect Speech Transcription

## 2.1 Text pre-processing

The text transcript is of limited size; an example is shown for the word category "significant" in Fig. 1. The speech transcript normalization is the process of cleaning the data by removing unwanted information- stop words, punctuation, conversion of numeral values to their word form, converting all words to lower case for better reading. Tokenizing of the sentence is performed to read the content properly and distinguish each word separately. Text filtering helps in faster processing and reduces the size of the document.

significance significant null null a significant significance vice president significance significant significant its significance i am significant significance significant significant significant significant significant significant significant significant i have significant significance significance more significant null null significant sunderkand significant

**Fig. 1.** Text transcription after pre-processing.

Stop words are removed since these have less importance, like “a”, “the”, “an”, “of”, “like”, “for” that are not of any significance in information retrieval. We have re-phrased commonly used phrases (like “couldn’t”) with their grammatical form (“could not”). Tokens containing symbols like “.”, “!”, “#”, “$” are converted to word form as per the requirement of the content, are removed, as shown in Fig. 2.

significance significant **null null a** significant significance vice president significance significant significant **its** significance **i am** significant significance significant significant significant significant significant significant significant significant **i have** significant significance significance **more** significant **null null** significant sunderkand significant

**Fig. 2.** After text normalization (removing all these bold letter words) from a transcription.

## 2.2 Stemming

In information retrieval and linguistic morphology, stemming is the process of chopping-off suffix from a word and reducing it to its root form also known as the base form. Stemming is used in text and natural language processing. For instance, if the word ends with “ed”, “ies”, “ing”, “ly” the end of the word is removed to obtain its root form or base form. One of the tools used for stemming is Porter stemmer where the stem may not be identical to the morphological root of the words, as observed from the stemmed results in Fig.3. Stemmer algorithm outlines the process of removing inflectional endings and common morphs from words. It is also used for text normalization in information retrieval systems (Porter, 1980) [7]. Porter stemmer is less aggressive than Lancaster stemmer which trims out most of the valid text. In linguistic morphology, stemming is the process of finding the root or base of the word. However, lemmatization is to find the lemma in a set of lexicons having same word sense. The base or word-form can be derived or inflected. In our LRW dataset, the keyword “absolutely” has lemma as “absolutely” and stem as “absolut”. When you go through the text transcription, the occurrence of stem “absolut” is more as compared to the lemma “absolutely”. So, the stemmer brings most of the words closer to their respective categories, that would yield higher classification scores than experiments that involve matching words in their original form as in [36] and [37].

signific signific signific signific vice presid signific signific signific signific signific signific signific signific signific signific signific signific signific signific signific signific signific signific signific sunderkand signific

**Fig. 3.** After applying stemming using Porter stemmer.

## 2.3 Phoneme Extraction from text

Phoneme is the smallest unit of speech in linguistics where a sound or group of sounds differ in their pronunciation and meaning, where the pronunciation varies according to the surrounding letters which may affect the letter representation [12]. The word can be sound of speech, including stress, articulation and intonation, for the representation of which we used CMU pronunciation dictionary with over 125k words and their phonetic transcriptions After text normalization, we generated a collection of phonemes from the text aligned transcript. Only the standard detailing pattern of the sound or stress pattern of a syllable, word, and phrase is selected. The phonetic transcription of the word “about” is “AH0 B AW1 T” where 0 represents no stress, 1 represents primary stress and 2 represents secondary stress; all the numerical values are filtered out. Thus the text is first segmented to words and further represented by their phonemes as shown in the example in Fig. 4.

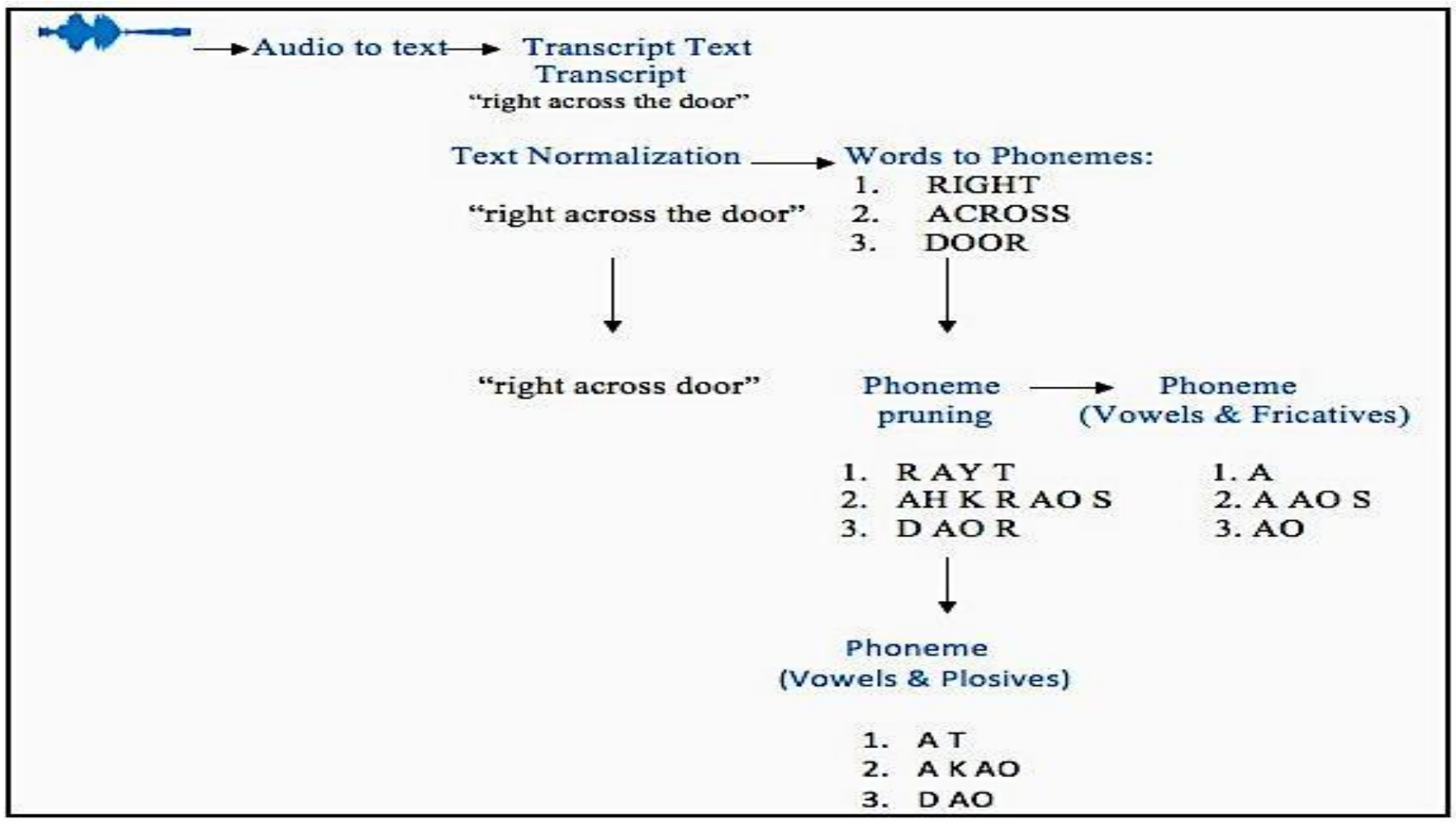

**Fig. 4.** Overview of phoneme filtering and pruning from a sample phrase. Text phoneme is filtered and pruned to include plosives and vowels and alternatively, to include vowels and fricatives in the same manner.

Different speakers contribute to a different pronunciation; searching for a perfect match in transcripts is challenging [22]. CMU pronouncing dictionary follows American English to assume pronunciation of a text. CMU pronouncing dictionary follows the same standard representation as the International phonetic alphabet of sounds in spoken language [40].

Phoneme filtering is the process where the phonemes are filtered on the basis of whether they contain vowels, plosives or fricatives. Phoneme filtering not only reduces the size of the dataset but also improves the category identification as shown from the sequence of phonemes shown in Fig. 5 that are extracted from the normalized text in Fig. 2. Plosives are also known as stop or oral consonant which block the vocal tract so that the flow of air ceases. They include both voiced and voiceless consonants. “b”, “d”, “g” are voiced plosives and “p”, “t”, “k” are voiceless or unvoiced plosives. They are also called glottal stop. Fricatives are consonants that are mostly voiced and consist of high energy and amplitude. The example of fricatives is “f”, “s”, “v”, “z”. Because of high amplitude they are easily detected. The International Phonetic alphabet (IPA) aims to transcribe the sounds of all human languages.

The IPA introduced the phonetic chart where each lexical unit is distinguished in the way it is spoken. Consonants are the sounds produced in the vocal tract, more specifically in the oral tract (the mouth and pharynx), where the produced speech is constructed. They are further classified to labials, coronal, radical and dorsal. The vowels contribute to high pitch and amplitude so most of the time they are detected correctly. Based on the manner the speech is formed, if it is sound coming from nose its called as nasal; if it is the sound formed by blocked air flow, then it is called stop, plosive, or oral. Fricatives are consonants such as (“f”, “s”, “v”, “z”) produced by placing the lower lip against the upper teeth. Trills are similar to taps and flaps is a sound produced by active and passive articulator (Richard and Makhoul,1975) [27].

S AH G N IH F IH K AH N S S IH G N IH F IH K AH N T S IH G N IH F IH K AH N T S
AH G N IH F IH K AH N S V AY S P R EH Z AH D AH N T S AH G N IH F IH K AH N
S S IH G N IH F IH K AH N T S IH G N IH F IH K AH N T S AH G N IH F IH K AH N S
S IH G N IH F IH K AH N T S AH G N IH F IH K AH N S S IH G N IH F IH K AH N T S
AH G N IH F IH K AH T S AH G N IH F IH K AH N T S AH G N IH F IH K AH N T S
AH G N IH F IH K AH N T S AH G N IH F IH K AH N T S AH G N IH F IH K AH N T S
AH G N IH F IH K AH N T S AH G N IH F IH K AH N T S AH G N IH F IH K AH N S S
AH G N IH F IH K AH N S S IH G N IH F IH K AH N T S IH G N IH F IH K AH N T S
AH N D ER K AH N D S IH G N IH F IH K AH N T

**Fig. 5.** After phoneme extraction and filtering.

## 2.4 Decision fusion of stemming and phoneme

After collecting results of pure stemming and phoneme filtering, we pruned each phoneme such that it contains either (i) vowels or plosives or (ii) vowels or fricatives. The pruning was performed in two non-sequential stages. We call the scheme two-way phoneme pruning.

Stage I: Phoneme pruning using Vowels and Plosives
Stage II: Phoneme pruning using Vowels and Fricatives

In order to combine the goodness of all factors, the outcomes of stemming and Stage I and Stage II of two-way phoneme pruning are combined using a decision level score fusion, by which scheme, if any of the stages (stem, vowel + plosives, vowel + fricatives) identify a given word in the transcript, then the word is considered to be identified.

# 3 Experimental Results

## 3.1 Dataset

One of the challenging datasets for speech recognition in the wild is Lip Reading in the wild (LRW) dataset [20]. It is an audio-visual dataset that has motivated various researches in audio-visual speech recognition [9, 21]. In our work, the audio track is extracted and processed for generating the speech transcription. The LRW dataset consists of 500 different classes of words (each class contains 50 samples). We use the testing data alone for the unsupervised experiments. All these videos are in MP4 format and have 29 frames each that are 1.16 seconds in length, and the word is supposed to occur in the middle of the video. The word length, word details are given in the metadata. To extract audio from the video we used Ffmpeg framework. Ffmpeg is a fast video and audio converter. There is no quality loss while changing the format of multimedia files.

## 3.2 Results of the proposed approach using decision-level fusion

The experiments were performed in python 3.7.4 on a PC (Intel core i5 with Intel HD Graphics 6000 1536 MB and Mac OS High Sierra with a clock of 1.8 GHz). The computation time for a single audio file is 29 seconds. In our proposed work, we filtered and pruned the phonemes by selecting vowels and plosives (in Stage 1) and vowels and fricatives (in Stage 2). After applying stemming, the word recognition rate was observed to be 23.34%. Stage 1 of Phoneme pruning yielded a recognition rate of 27.67% while Stage 2 yielded 28.23%. The results of decision-level fusion of stemming and Stage-1 and Stage-2 of phoneme pruning, yielded the word recognition rate of 32.96%, a significant improvement over the baseline method (9.36%) which is explained in the next sub-section. A demonstration of decision-level fusion is shown in Fig. 6 for the word category "significant". All results are summarized in Table 1 with suitable references to the literature regarding the use of the individual components for speech text understanding. We have also compared our approach with two well- known automated spelling correction tools – autocorrect [30] and symspell [6]. The code for both these tools is available online at [32] and [19] respectively.

| a) Phoneme pruning using vowels and plosives | b) Phoneme pruning using vowels and fricatives | c) Stemming |
|---|---|---|
| A G I I K A | S A I F I A S | signific |
| A G I I K A T | S A I F I A | signific |
| <unk> | <unk> | skiwrd |
| <unk> | <unk> | signuyt |
| A G I I K A T | S A I F I A | signific |
| A G I I K A | S A I F I A S | signific |
| A P E A D E T | V A S E Z A E | vice presid |
| A G I I K A | S A I F I A S | signific |
| A G I I K A T | S A I F I A | signific |
| A G I I K A T | S A I F I A | signific |
| A G I I K A | S A I F I A S | signific |
| A G I I K A T | S A I F I A | signific |
| A G I I K A | S A I F I A S | signific |
| A G I I K A T | S A I F I A | signific |
| A G I I K A T | S A I F I A | signific |
| A G I I K A T | S A I F I A | signific |
| A G I I K A T | S A I F I A | signific |
| A G I I K A T | S A I F I A | signific |
| A G I I K A T | S A I F I A | signific |
| A G I I K A T | S A I F I A | signific |
| A G I I K A T | S A I F I A | signific |
| Number of significant: 13 | Number of significant: 5 | Number of significant: 19 |

After Applying Decision Fusion on Stemming and Phonemes: 1 1 0 0 1 1 0 1 1 1 1 1 1 1 1 1 1 1 1
Occurrence of stem significant: 19
Probability of occurrence of word significant after fusion: 86.36%

**Fig. 6.** Decision fusion of stemming and two-way phoneme pruning.

**Table 1.** Comparison of various methodologies on LRW dataset.

| Methods | Accuracy |
|---|---|
| Baseline [29] | **9.34%** |
| Stemming [28] | **23.34%** |
| Phoneme pruning (Vowels & Plosives) [3] | **27.67%** |
| Phoneme pruning (Vowels & Fricatives) [5] | **28.23%** |
| Autocorrect [30] | **21.50%** |
| Symspell [6] | **25.16%** |
| Decision Fusion of Stemming and two-way Phoneme pruning | **32.96%** |

### 3.3 Sliding Text window-The baseline approach

In the baseline approach, we slide a window across the text after tokenizing each word in the sentence. If the category word is found in a sentence, the text window can slide to the next sentence or next line in the text file. After sliding word by word in a sentence and line by line, it determines the number of occurrences of the category word, as shown in Fig. 7. In this way the duplicity and redundancy of the same category in a sentence can be avoided.

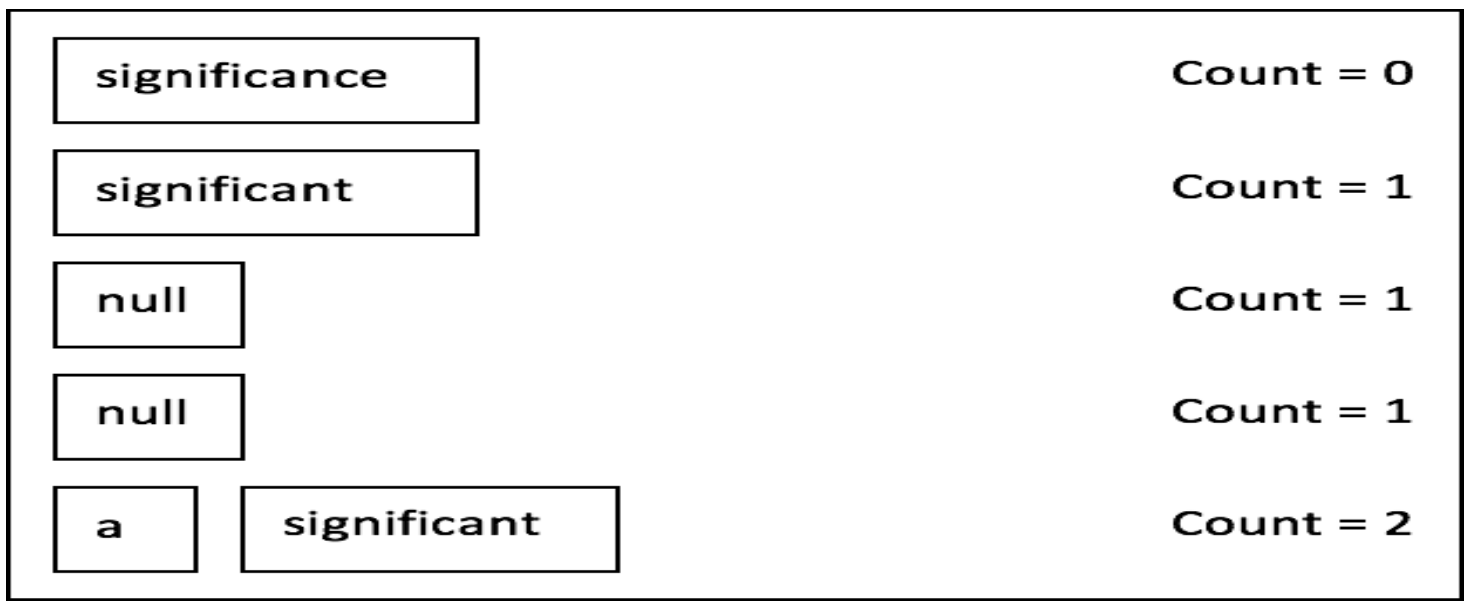

**Fig. 7.** Sliding text window to search for the keyword “significant” in-text transcription.

### 3.4 Test cases

All 500 categories results (averaged) are projected in Table 1. An examination of test cases shows that sometimes stemming performs best, sometimes phonemes (vowels + plosives) performs best, and sometimes phonemes (vowels + fricatives) gives the best results. Our technique combines the best of all three methods and provides the highest accuracy. Some test cases are presented below, that can help us in understanding the role of decision fusion more precisely.

**Test case 1:** In some cases, only stem is found sufficient to recognize the word. For instance, in category "announced," the probability of occurrence of the word "announce", which is the root form of the word "announced", is 6%, the probability of occurrence of vowels + plosives is 4%, and the probability of occurrence of vowels + fricatives is 2%. As you can see, in such cases, decision fusion causes the stem to win with a high probability value. After trimming “ed” from the word “announced”, the probability of occurrence of stem “announce” increases by 4%. However, in vowels + plosives, the word “announced” is pronounced in CMU dictionary as “AH”, “N”, “AW”, “N”, “S”, “T”. The vowels are- “A” and plosives- “T”, the pattern we are looking for in our transcript is “AAT”; the probability of increase in this case is 2%. Moreover, in case of vowels + fricatives, the word “announced” is pronounced in CMU dictionary as “AH”, “N”, “AW”, “N”, “S”, “T’. The vowel is- “A” and fricatives are- “W”, “S”. The pattern we are looking for in our transcript is “AAWS”. The probability of occurrence in this case is same as the baseline. The probability of increase is more in case of stem than other two cases. Hence, stemming performs best for this particular test case.

**Test case 2:** Let’s take one more example of another category “agreement” where the probability of occurrence of stem is 44%, the probability of occurrence of vowels + plosives is 88% and the probability of occurrence of vowel + fricatives is 92%. In this case, decision fusion causes vowels + fricatives to win with high probability value. In category “agreement”, the probability of occurrence of stem “agreement” is same as the baseline. However, in vowels + plosives, the word “agreement” is pronounced in CMU dictionary as “AH”, “G”, “R”, “IY”, “M”, “AH”, “N”, “T”. The vowels are- “A”, “I” and plosives are- “G”, “T”. The pattern we are looking for in our transcript is “AGIAT”, the probability of increase in this case is 22%. However, in case of vowels + fricatives, the word “agreement” is pronounced in CMU dictionary as “AH”, “G”, “R”, “IY”, “M”, “AH”, “N”, “T”. The vowels is- “A”, “I” and plosives are- none the pattern we are looking for in our transcript is “AI”. The probability of increase in this case is 70%. The probability of increase is more in case of vowels + fricatives than the other two cases. Hence, vowels + fricatives perform best in this test case.

**Test case 3:** In category “affairs” the probability of occurrence of stem is 4 %, the probability of occurrence of vowels + plosives is 20%, and the probability of occurrence of vowels + fricatives is 6%. This is the case in which decision fusion causes vowels + plosives to win with high probability value. After trimming “s” from the word “affairs”, the probability of occurrence of stem “affair” is same as the baseline. However, in vowels + plosives, the word “affairs” is pronounced in CMU dictionary as “AH”, “F”, “EH”, “R”, “Z”. The vowels are- “A”, “E” and plosives are- none, the pattern we are looking for in our transcript is “AE”. The probability of increase in this case is 16%. However, in case of vowels + fricatives, the word “affairs” is pronounced in CMU dictionary as “AH”, “F”, “EH”, “R”, “Z”. The vowels are- “A”, “E” and fricatives are- “F”, “Z”, the pattern we are looking for in our transcript is “AFERZ”. The probability of increase in this case is 2%. The probability of increase is more in case of vowels + plosives than the other two cases. Hence, vowels + plosives give the best results for this particular test case.

## 4 Conclusions

We have presented a fusion approach of the best of stemming and two-way phoneme pruning on highly imperfect speech transcription extracted from the LRW dataset which is in mp4 format. After extracting audio samples using Ffmpeg framework, we converted the audio speech to the text transcription using Google API which is publicly available and has various applications in speech adaption, transcribing speech and real-time speech recognition. We have evaluated the baseline results by pure string matching of a word category from a text transcription. The first step is text normalization and speech adaptation by removing stop words which are the most frequent unwanted words from a text file to make text processing faster. After applying stemming on the word, we extracted the most root word and compared it with different categories. At the same time, we converted the word to phonemes using the CMU pronouncing dictionary. After mapping text transcript to phonemes, we applied phoneme filtering on text transcript, where we filtered out the phonemes containing vowels, plosives or fricatives. The phoneme pruning is executed in two non-sequential stages: Stage I: Phoneme pruning using Vowels and Plosives, Stage II: Phoneme pruning using Vowels and Fricatives. Once we got results through the above three methods, we applied decision fusion that confirmed whether the occurrence of the word is detected by any of the three methods. The proposed fusion method outperforms the state of the art and the word recognition accuracy is improved from the baseline accuracy of 9.34% to 32.96% using our fusion method.

## References

1. Besacier, Laurent, Etienne Barnard, Alexey Karpov, and Tanja Schultz. "Automatic speech recognition for under-resourced languages: A survey." *Speech Communication* 56 (2014): 85-100.
2. Susan, Seba, and Srishti Sharma. "A fuzzy nearest neighbor classifier for speaker identification." In *2012 Fourth International Conference on Computational Intelligence and Communication Networks*, pp. 842-845. IEEE, 2012.
3. Hemakumar, G. "Vowel-Plosive of English Word Recognition using HMM." *IJCSI* (2011).
4. Tripathi, Mayank, Divyanshu Singh, and Seba Susan. "Speaker Recognition using SincNet and X-Vector Fusion." *arXiv preprint arXiv:2004.02219* (2020).
5. Haubold, Alexander, and John R. Kender. "Alignment of speech to highly imperfect text transcriptions." In *2007 IEEE International Conference on Multimedia and Expo*, pp. 224-227. IEEE, 2007.
6. Gupta, Prabhakar. "A context-sensitive real-time Spell Checker with language adaptability." In *2020 IEEE 14th International Conference on Semantic Computing (ICSC)*, pp. 116-122. IEEE, 2020.
7. Porter, Martin F. "An algorithm for suffix stripping." *Program* 14, no. 3 (1980): 130-137.
8. Stan, Adriana, Peter Bell, and Simon King. "A grapheme-based method for automatic alignment of speech and text data." In *2012 IEEE Spoken Language Technology Workshop (SLT)*, pp. 286-290. IEEE, 2012.
9. Haubold, Alexander, and John R. Kender. "Augmented segmentation and visualization for presentation videos." In *Proceedings of the 13th annual ACM international conference on Multimedia*, pp. 51-60. 2005.
10. Ghosh, Krishnendu, and K. Sreenivasa Rao. "Subword based approach for grapheme-to-phoneme conversion in Bengali text-to-speech synthesis system." In *2012 National Conference on Communications (NCC)*, pp. 1-5. IEEE, 2012.
11. Wang, Weiran, Yingbo Zhou, Caiming Xiong, and Richard Socher. "An investigation of phone-based subword units for end-to-end speech recognition." *arXiv preprint arXiv:2004.04290* (2020).
12. Alsharhan, Eiman, and Allan Ramsay. "Improved Arabic speech recognition system through the automatic generation of fine-grained phonetic transcriptions." *Information Processing & Management* 56, no. 2 (2019): 343-353.
13. Gimenes, Manuel, Cyril Perret, and Boris New. "Lexique-Infra: grapheme-phoneme, phoneme-grapheme regularity, consistency, and other sublexical statistics for 137,717 polysyllabic French words." *Behavior Research Methods* (2020).
14. Harwath, David, and James Glass. "Towards visually grounded sub-word speech unit discovery." In *ICASSP 2019-2019 IEEE International Conference on Acoustics, Speech and Signal Processing (ICASSP)*, pp. 3017-3021. IEEE, 2019.
15. Lin, Shih-Hsiang, Yao-Ming Yeh, and Berlin Chen. "Extractive speech summarization-From the view of decision theory." In *Eleventh Annual Conference of the International Speech Communication Association*. 2010.
16. Siivola, Vesa, Teemu Hirsimaki, Mathias Creutz, and Mikko Kurimo. "Unlimited vocabulary speech recognition based on morphs discovered in an unsupervised manner." In *Eighth European Conference on Speech Communication and Technology*. 2003.
17. Williamson, Donald S., Yuxuan Wang, and DeLiang Wang. "Complex ratio masking for monaural speech separation." *IEEE/ACM transactions on audio, speech, and language processing* 24, no. 3 (2015): 483-492.

18. Chen, Jitong, Yuxuan Wang, and DeLiang Wang. "A feature study for classification-based speech separation at low signal-to-noise ratios." *IEEE/ACM Transactions on Audio, Speech, and Language Processing* 22, no. 12 (2014): 1993-2002.
19. Mamamothb,"Python port SymSpell", 2019. [Online]. https://github.com/mammothb/symspellpy
20. Yang, Shuang, Yuanhang Zhang, Dalu Feng, Mingmin Yang, Chenhao Wang, Jingyun Xiao, Keyu Long, Shiguang Shan, and Xilin Chen. "LRW-1000: A naturally-distributed large-scale benchmark for lip reading in the wild." In *2019 14th IEEE International Conference on Automatic Face & Gesture Recognition (FG 2019)*, pp. 1-8. IEEE, 2019.
21. Torfi, Amirsina, Seyed Mehdi Iranmanesh, Nasser Nasrabadi, and Jeremy Dawson. "3d convolutional neural networks for cross audio-visual matching recognition." *IEEE Access* 5 (2017): 22081-22091.
22. Hazen, Timothy J. "Automatic alignment and error correction of human generated transcripts for long speech recordings." In *Ninth International Conference on Spoken Language Processing*. 2006.
23. Martin, Philippe. "WinPitchPro-A Tool for Text to Speech Alignment and Prosodic Analysis." In *Speech Prosody 2004, International Conference*. 2004.
24. Chen, Yi-Chen, Chia-Hao Shen, Sung-Feng Huang, and Hung-yi Lee. "Towards unsupervised automatic speech recognition trained by unaligned speech and text only." *arXiv preprint arXiv:1803.10952* (2018).
25. Novotney, Scott, Richard Schwartz, and Jeff Ma. "Unsupervised acoustic and language model training with small amounts of labelled data." In *2009 IEEE International Conference on Acoustics, Speech and Signal Processing*, pp. 4297-4300. IEEE, 2009.
26. https:// github.com/wolfgarbe/SymSpell
27. Schwartz, Richard, and John Makhoul. "Where the phonemes are: Dealing with ambiguity in acoustic-phonetic recognition." *IEEE Transactions on Acoustics, Speech, and Signal Processing* 23, no. 1 (1975): 50-53.
28. Mulholland, Matthew, Melissa Lopez, Keelan Evanini, Anastassia Loukina, and Yao Qian. "A comparison of ASR and human errors for transcription of non-native spontaneous speech." In *2016 IEEE International Conference on Acoustics, Speech and Signal Processing (ICASSP)*, pp. 5855-5859. IEEE, 2016.
29. Bahl, L., S. Das, P. de Souza, Frederick Jelinek, S. Katz, R. Mercer, and M. Picheny. "Some experiments with large-vocabulary isolated-word sentence recognition." In *ICASSP'84. IEEE International Conference on Acoustics, Speech, and Signal Processing*, vol. 9, pp. 395-396. IEEE, 1984.
30. Rayson, Steven J., Dean J. Hachamovitch, Andrew L. Kwatinetz, and Stephen M. Hirsch. "Autocorrecting text typed into a word processing document." U.S. Patent 5,761,689, issued June 2, 1998.
31. Xu, Hainan, Shuoyang Ding, and Shinji Watanabe. "Improving end-to-end speech recognition with pronunciation-assisted sub-word modeling." In *ICASSP 2019-2019 IEEE International Conference on Acoustics, Speech and Signal Processing (ICASSP)*, pp. 7110-7114. IEEE, 2019.
32. https://github.com/phatpiglet/autocorrect
33. Drexler, Jennifer, and James Glass. "Learning a Subword Inventory Jointly with End-to-End Automatic Speech Recognition." In *ICASSP 2020-2020 IEEE International Conference on Acoustics, Speech and Signal Processing (ICASSP)*, pp. 6439-6443. IEEE, 2020.
34. Hermann, Enno, Herman Kamper, and Sharon Goldwater. "Multilingual and unsupervised subword modeling for zero-resource languages." *Computer Speech & Language* (2020): 101098.

35. Agenbag, Wiehan, and Thomas Niesler. "Automatic sub-word unit discovery and pronunciation lexicon induction for ASR with application to under-resourced languages." *Computer Speech & Language* 57 (2019): 20-40.
36. Susan, Seba, Shashank Kumar, Rohen Agrawal, and Kartik Yadav. "Statistical keyword matching using Automata." *International Journal of Applied Research on Information Technology and Computing* 5, no. 3 (2014): 250-255.
37. Susan, Seba, and Juli Keshari. "Finding significant keywords for document databases by two-phase Maximum Entropy Partitioning." *Pattern Recognition Letters* 125 (2019): 195-205.
38. Feng, Siyuan, and Tan Lee. "Exploiting cross-lingual speaker and phonetic diversity for unsupervised subword modeling." *IEEE/ACM Transactions on Audio, Speech, and Language Processing* 27, no. 12 (2019): 2000-2011.
39. Ojha, Rupam, and C. Chandra Sekhar. "Multi-label Classification Models for Detection of Phonetic Features in building Acoustic Models." In *2019 International Joint Conference on Neural Networks (IJCNN)*, pp. 1-8. IEEE, 2019.
40. CMU Pronouncing Dictionary: www.speech.cs.cmu.edu/cgi-bin/cmudict [Last accessed online on 15th June 2020].